\documentclass[12pt]{article}

\newtheorem{theorem}{Theorem}
\newtheorem{corollary}{Corollary}
\newtheorem{lemma}{Lemma}

\usepackage{times,epsfig,graphics,amssymb,latexsym,amsmath,setspace,fullpage,color}
\usepackage{array,threeparttable}

\def\argmax{\mathop{\rm argmax}}
\def\argmin{\mathop{\rm argmin}}

\def\Cov{\mathop{\rm Cov}}

\def\sign{\mathop{\rm sign}}

\begin{document}
\pagenumbering{arabic}

\setcounter{page}{1}

\title{\Large \bf An efficient model-free estimation of multiclass conditional probability}
\author{
    Tu Xu and Junhui Wang\\
    Department of Mathematics, Statistics, \\
    and Computer Science \\
    University of Illinois at Chicago \\
    Chicago, IL 60607
}
\date{}
\maketitle

\begin{abstract} {\noindent Conventional multiclass conditional probability estimation methods, such as Fisher's discriminate analysis and logistic regression, often require restrictive distributional model assumption. In this paper, a model-free estimation method is proposed to estimate multiclass conditional probability through a series of conditional quantile regression functions. Specifically, the conditional class probability is formulated as difference of corresponding cumulative distribution functions, where the cumulative distribution functions can be converted from the estimated conditional quantile regression functions.  The proposed estimation method is also efficient as its computation cost does not increase exponentially with the number of classes. The theoretical and numerical studies demonstrate that the proposed estimation method is highly competitive against the existing competitors, especially when the number of classes is relatively large. }
\end{abstract}


\doublespacing

\section{Introduction}

Estimation of conditional class probability is important in statistical machine learning since the conditional class probability measures the strength and confidence of the classification outcomes. It also provides supplemental information to the classification labels, such as hazard reduction in ``evidence-based'' medication (Wahba, 2002) and pixel spectrum in remote sensing (Xu, 2005). In multiclass classification, a training sample $\{({\bf x}_i,y_i); i=1,2,\ldots,n\}$ is available with covariate ${\bf x}_i \in {\cal R}^p$ and class label $y_i \in \{1,2,\ldots,K\}$, where $K$ is the number of classes. Due to the discrete feature, the conditional distribution of $Y$ given ${\bf X}={\bf x}$ can be fully characterized by the conditional class probability $p_k({\bf x})=P(Y=k|{\bf X}={\bf x})$. Estimation of $p_k({\bf x})$ is the primary goal of this paper, which is also known as the soft classification (Wahba, 2002; Liu, Zhang and Wu, 2011), as opposed to the hard classification that mainly focuses on predicting the class labels without estimating probability. 

In literature, many classical probability estimation methods have been developed based on certain distributional model assumptions. For
instances, Fisher's discriminant analysis assumes that the
covariates within each class follow multivariate
Gaussian distributions with homogeneous or heteroscedastic covariance
matrices. Relaxing the Gaussian distribution assumption, the
multiple logistic regression takes one class as baseline and
assumes the logarithms of all the odds ratios are linear functions of the covariates. Although these estimation methods have been widely used in practice, it is generally difficult to verify the distributional model assumptions and thus may lead to suboptimal performance when the assumptions are violated.  

To circumvent the restrictive distributional assumption, various model-free probability estimation methods have been proposed and gained their popularity among the practitioners. Classification tree is a popular model-free classification method that produces probabilistic outputs, however it can be over-sensitive to the training set and thus suffers from issues of over-fitting and instability (Breiman, 1996). Wang, Shen and Liu (2008) proposes a model-free binary conditional probability estimation method by bracketing the conditional probability through a
series of weighted binary large-margin classifiers with various
weights $\pi \in (0,1)$. The method is based on the property that the
consistent weighted binary large-margin classifiers aim at
estimating $\sign(p_1({\bf x})-\pi)$, and hence that the small bracket $(\pi,\pi')$ containing $p_1({\bf x})$ can be obtained based on the estimated $\sign(p_1({\bf x})-\pi)$ for different $\pi$'s. To extend the binary estimation method to multiclass case, a number of attempts have been proposed. Hastie and Tibshirani (1998) and Wu, Lin and Weng (2004) develop the pairwise
coupling method, which converts the multiclass probability estimation
into estimating multiple one-vs-one binary conditional probabilities. Wu, Zhang and Liu (2010) directly extends the idea of Wang, Shen and Liu (2008) and designs an interesting way of assigning weights to the multiple classes, and then produce the estimated conditional probability by searching for the $K$-vertex polyhedron that contains $p_k({\bf x})$. However, both methods require intensive
computational cost as the number of one-vs-one binary classifications is
proportional to $K^2$ and the number of $K$-vertex polyhedrons
increases with $K$ exponentially.

In this paper, an efficient bracketing scheme is proposed for
estimating the multiclass conditional probability via a series of estimated
conditional quantile functions (Koenker and Bassett, 1978; Koenker, 2005). The key idea is that $p_k({\bf x})$ can be formulated as the difference of corresponding cumulative distribution functions $P(Y \leq k|{\bf X}={\bf
x})$, which can be obtained through a series of estimated conditional quantiles of $Y$ given ${\bf X}={\bf x}$. Compared with other model-free estimation methods, the proposed estimation method is computationally efficient in that its computational cost does not increase with $K$ exponentially, which is desirable especially when $K$ is large. The solution surface of the regularized quantile regression estimation (Rosset, 2009) can further alleviate the computation burden. More importantly, the asymptotic property of the proposed estimation method is established, which shows that the proposed estimation method achieves a fast convergence rate to the true $p_k({\bf x})$. The simulation studies and real data analysis also demonstrate that the proposed method is highly competitive against the existing competitors.

The rest of the paper is organized as follows. Section 2 presents the proposed multiclass conditional probability estimation method along with its computational implementation. A tuning parameter selection criterion is also introduced. Section 3 establishes the asymptotic convergence property of the proposed method. Section 4 examines the numerical performance of the proposed estimation method in both simulated examples and real applications. Section 5 contains some discussion, and the appendix is devoted to technical proofs.

\section{Multiclass probability estimation via quantile estimation}
This section presents the novel model-free estimation method for multiclass conditional probability and its computational implementation.

\subsection{Multiclass probability estimation via quantile regression}
In multiclass classification with $Y \in \{1, \ldots, K\}$, estimation of $p_k({\bf x})$ is equivalent to estimation of $P(Y \leq k|{\bf X}={\bf x})$ due to the following decomposition,
\begin{equation}
\label{prob}
p_k({\bf x})=P(Y \leq k|{\bf X}={\bf x})-P(Y \leq k-1|{\bf X}={\bf x}),
\end{equation}
where $P(Y \leq k|{\bf X}={\bf x})=\sum_{j=1}^k p_j({\bf x})$ is the
conditional cumulative distribution function of $Y$ given ${\bf X}={\bf x}$. Furthermore, the estimated $P(Y \leq k|{\bf X}={\bf x})$ can be constructed through a series of estimated quantile regression functions, since
\begin{equation}
\label{eqn:connect}
P(Y\leq k|{\bf X}={\bf x})=\argmax_{\tau}~\{f^*_{\tau}({\bf x}) \leq k\},
\end{equation}
where $f^*_{\tau}({\bf x})$ represents the $\tau$-th conditional quantile of $Y$ given ${\bf X = x}$, defined as
$$
f^*_\tau({\bf x})=\argmin_y~\{y: P(Y \leq y|{\bf X}={\bf x}) \geq \tau \}.
$$

Since $Y$ is discrete and only takes values in $\{1,\cdots,
K\}$, estimating $f^*_\tau({\bf x})$ can encounter various difficulties such as discontinuity as discussed in Machado et al. (2005) and Chen et al. (2010). A simple treatment is to jitter the discrete response by adding some continuous noises. In specific, denote the jittered response
$\tilde{Y}=Y+\epsilon$, where $\epsilon$ follows a uniform
distribution on $(-0.5,0.5)$ and is independent of $Y$, and denote
$\tilde{f}^*_\tau({\bf x})$ as the $\tau$-th quantile of $\tilde{Y}$
given ${\bf X}={\bf x}$. With jittering, $\tilde{Y}$ becomes continuous, $P(\tilde{Y} \leq y)$ is strictly increasing in $y$, and thus $\tilde{f}^*_\tau({\bf x})$ is also continuous and strictly increasing in $\tau$. More
importantly, $P(Y \leq k)=P(\tilde Y \leq k+0.5)$, and
$$
f^*_\tau({\bf x})=k~~\mbox{if and only if}~~\tilde{f}^*_\tau({\bf x}) \in (k-0.5, k+0.5).
$$
Combining the results, $\tilde{f}^*_\tau({\bf x})$ can be explicitly connected with $p_k({\bf x})$ as in Lemma 1.

{\lemma The $\tau$-th quantile of $\ \tilde{Y}=Y+\varepsilon$ given
${\bf X}={\bf x}$ is
\begin{equation}
\label{quantile}
\tilde{f}^*_{\tau}({\bf x}) =
k-0.5+\frac{\tau-\sum\limits_{j=0}^{k-1} p_j({\bf x})} {p_k({\bf
x})},~~\mbox{if}~~ \sum\limits_{j=0}^{k-1} p_j({\bf x})< \tau \leq
\sum\limits_{j=0}^k p_j({\bf x}),
\end{equation}
where $p_0({\bf x})$ is set to be 0 for simplicity. }

By Lemma 1, $P(\tilde{Y}\leq k+0.5|{\bf X}={\bf x})=
\argmax\limits_{\tau}~\{ \tilde{f}^*_{\tau}({\bf x}) \leq k+0.5\}$, and then
\begin{eqnarray}
p_k (x) &=& P(\tilde{Y} \leq k+0.5|{\bf X}={\bf x})-P(\tilde{Y} \leq
k-0.5|{\bf X}={\bf x}) \nonumber  \\
&=& \argmax_{\tau}~\{ \tilde{f}^*_{\tau}({\bf x}) \leq k+0.5\} - \argmax_{\tau}~\{ \tilde{f}^*_{\tau}({\bf x}) \leq k-0.5\}.
\label{scheme}
\end{eqnarray}
Therefore, estimation of $p_k({\bf x})$ boils down to
estimating quantile regression function
$\tilde{f}^*_\tau({\bf x})$ for various $\tau$'s. Specifically, let $0=\tau_0<\tau_1<\tau_2<\cdots<\tau_{m-1}<\tau_m=1$ be a sequence of $\tau$'s, and $\hat{f}_{\tau_1}({\bf x}),\hat{f}_{\tau_2}({\bf x}),\ldots$,
$\hat{f}_{\tau_{m-1}}({\bf x})$ be the estimated $\tilde{f}^*_\tau({\bf x})$'s. According to (\ref{scheme}), $p_k({\bf x})$ can be estimated as
\begin{equation}
\label{scale1}
\hat{p}_k({\bf x})=\argmax_{\tau_j}~\{ \hat{f}_{\tau_j}({\bf x}) \leq k+0.5\}
- \argmax_{\tau_j}~\{ \hat{f}_{\tau_j}({\bf x}) \leq k-0.5\},
\end{equation}
where $\hat{f}_{\tau_0}({\bf x})=0.5$ and $\hat{f}_{\tau_m}({\bf x})=K+0.5$ for simplicity.

Note that $\hat{f}_{\tau}({\bf x})$ can be estimated by any existing quantile regression estimation method, such as He, Ng and Portnoy (1998), Li, Liu and Zhu (2007), Wang, Zhu and Zhou (2009), Yang and He (2012), and many others. For illustration, we adopt the nonparametric method in Li, Liu and Zhu (2007), which is formulated as

\begin{equation}
\label{main} \min_{f_\tau \in {\cal H}_K} \sum\limits_{i=1}^n
\rho_\tau (y_i-f_\tau ({\bf x}_i)) + \frac{\lambda}{2} \|f_\tau\|^2_{{\cal
H}_K},
\end{equation}
where ${\cal H}_K$ is a reproducing kernel Hilbert spaces (RKHS; Wahba 1990) induced by a pre-specified kernel function $K(\cdot,\cdot)$, $\rho_\tau$ is the check loss function and $J(f_\tau)=\frac{1}{2}\| f_\tau \|_{{\cal H}_K}^2$ is the associated RKHS norm.
It is shown in Li, Liu, and Zhu (2007) that the estimated $\hat{f}_\tau ({\bf x})$ based on (\ref{main}) converges to $\tilde{f}^*_\tau({\bf x})$ in terms of $e(\hat{f}_\tau,\tilde{f}^*_\tau)=R(\hat{f}_\tau)-R(\tilde{f}^*_\tau)$ for any $\tau$, where $R(f_\tau)=E(\rho_\tau(Y-f_\tau({\bf X})))$.

As computational remarks, the proposed estimation method in (\ref{scale1}) only requires fitting $m-1$ conditional quantile functions. The optimal value of $m$, as shown in Section 3, only relies on the asymptotic behavior of the quantile regression estimation.
The grid points $\tau_1,\ldots,\tau_{m-1}$ can be simply set as equally spaced points on $(0,1)$, and more sophisticated adaptive design can be employed as well. For comparison, when the number of grid points along each direction is $m$, the computational complexity of the proposed method is $O(mn^3)$, whereas the complexity of the method in Wu et al. (2010) is $O(m^{K-1} n^3)$. It is clear that 
the proposed method is computationally more efficient as its complexity does not increase exponentially with $K$. Furthermore, although the true $\tilde f^*_\tau({\bf x})$ is strictly increasing in $\tau$, the
fitted quantile regression functions $\hat f_{\tau}({\bf x})$ may
cross each other and thus become inconsistent with order of $\tilde f^*_\tau({\bf
x})$ (He, 1997), leading to suboptimal estimation of $\hat p_k({\bf x})$ in practice. To prevent that from happening, some non-crossing constraints as in Wu and Liu (2009), Bondell, Reich and Wang (2010) and Liu and Wu (2011) can be enforced. Finally, the estimation performance of (\ref{main}) largely depends on the choice of
tuning parameter $\lambda$, which needs to
be appropriately determined.

\subsection{Model tuning and solution surface}

In this section, a data adaptive model tuning method for
multiclass conditional probability estimation is developed. To
indicate the dependency on the tuning parameter $\lambda$, we denote
the estimated conditional probability as $\hat p_\lambda({\bf x}) =
(\hat{p}_1({\bf x}), \ldots, \hat{p}_K({\bf x}))^T$ and the
quantile regression function as $\hat f_{\lambda,\tau}({\bf x})$.
The overall performance of $\hat{p}_\lambda({\bf x})$ in estimating
$p({\bf x})=(p_1({\bf x}),\ldots ,p_K({\bf x}))^T$ is evaluated by the generalized Kullback-Leibler (GKL) loss between $p$ and $\hat{p}_\lambda$,
\begin{equation}
GKL(p,\hat p_\lambda) = E \left ( \sum_{k=1}^K p_k({\bf X}) \log
\frac{p_k({\bf X})}{\hat p_k({\bf X})} \right ). \label{GKL}
\end{equation}
The corresponding comparative GKL loss, after omitting $\hat
p_\lambda$-unrelated terms in (\ref{GKL}), is
$$
GKL^c(p,\hat p_\lambda)=-\sum_{k=1}^K E( p_k({\bf X}) \log(\hat
p_k({\bf X}))).
$$
It is natural to estimate $GKL^c(p,\hat p_\lambda)$ by its empirical version,
\begin{equation}
EGKL(\hat{p}_\lambda) = - n^{-1} \sum_{k=1}^K \sum_{i=1}^n I(Y_i=k)
\log \hat p_k({\bf x}_i), \label{EGKL}
\end{equation}
where $I(\cdot)$ is an indicator function. However, $EGKL(\hat{p}_\lambda)$ often underestimates $GKL^c(p,\hat p_\lambda)$ especially when the estimation model is over-complicated.

To remedy the underestimation bias, $GKL^c(p,\hat p_\lambda)$ can be estimated similarly as in Wang, Shen and Liu (2008) by searching for the optimal correction terms for $EGKL(\hat{p}_\lambda)$. Specifically, minimizing the $L_2$ distance between $GKL^c(p,\hat p_\lambda)$ and a class of candidate estimators of form $EGKL(\hat{p}_\lambda) + {\bf X}^n$-dependent penalty with ${\bf X}^n=\{{\bf x}_i\}_{i=1}^n$ yields that
$$
\widehat{GKL}^c(p,\hat{p}_\lambda)=EGKL^c(\hat{p}_\lambda)+n^{-1}
\sum_{k=1}^K \sum_{i=1}^n \widehat{\Cov} \big (I(Y_i=k), \log(\hat
p_k({\bf x}_i)) |{\bf X}^n \big ) +
\widehat{D}_n(\hat{p}_\lambda,{\bf X}^n),
$$
where $D_n(\hat{p}_\lambda,{\bf X}^n)=\sum_{k=1}^K E \big (n^{-1}
\sum_{i=1}^n p_k({\bf x}_i) \log(\hat p_k({\bf x}_i)) - E(p_k({\bf
X}) \log(\hat p_k({\bf X})))|{\bf X}^n \big )$.
Here, $\Cov \big (I(Y_i=k), \log(\hat p_k({\bf x}_i)) |{\bf X}^n
\big )$ evaluates the accuracy of estimating $\hat{p}_k$ on ${\bf
X}^n$, which is similar to the covariance penalty in Efron (2004)
and the generalized degree of freedom in Shen and Huang (2006), and
the term $D_n(\hat{p}_\lambda,{\bf X}^n)$ is a correction term adjusting
the effect of random covariates ${\bf X}$ on prediction and needs to be
estimated, c.f., Breiman and Spector (1992), and Breiman (1992).

To construct the estimated $\widehat{\Cov}
\big (I(Y_i=k), \log(\hat p_k(x_i)) |{\bf X}^n \big )$ and $
\widehat{D}_n(\hat{ p}_\lambda,{\bf X}^n)$, the data
perturbation technique (Wang and Shen, 2006) can be adopted. The key idea is
to evaluate the generalization ability of the probability estimation
method by its sensitivity to the local perturbations of ${\bf X}$
and $Y$. The estimation formula can be derived via derivative
estimation and approximated through a Monte Carlo approximation.
The exact expressions are similar to (11) and (12) in
Wang, Shen and Liu (2007) and thus omitted here.

Note that the data perturbation technique requires fitting the
quantile regression function multiple times for various $\tau$'s and $\lambda$'s,
and thus can be computationally expensive. To further reduce the
computation cost, the solution surface of the coefficient of $\hat
f_{\lambda,\tau}({\bf x})$ with respect to $\lambda$ and $\tau$ can be
constructed following Rosset (2009). In particular, Li et al. (2007)
and Takeuchi et al. (2009) show that the solution path of $\hat
f_{\lambda,\tau}({\bf x})$ is piecewise linear with respect to $\lambda$ (or
$\tau$) when $\tau$ (or $\lambda$) is fixed; Rosset (2009) explores
the bi-level path of regularized quantile regression and shows that the solution surface of $\hat f_{\lambda,\tau}({\bf x})$ can be efficiently constructed with respect to both $\lambda$ and $\tau$. The solution surface is mapped as a piecewise linear function of $\tau$ or $\lambda$ and the possible locations of the bi-level optima can be found in one run of the base algorithm. That being said, the coefficient of $\hat f_{\lambda,\tau}({\bf x})$ for various $\lambda$'s and $\tau$'s can be obtained at essentially the same
computation cost as fitting one time of the base algorithm. Figure 1
displays $\hat{f}_{\tau,\lambda}({\bf x})$ for a fixed $\bf x$ as a function of $\lambda$ and $\tau$ in a randomly selected replication of the simulated Example 1.

\begin{center}
\begin{tabular}[t]{c}
      \hline
      \hline
      Figure 1 here\\
      \hline
      \hline
\end{tabular}
\end{center}

\section{Statistical learning theory}

This section establishes the asymptotic convergence of the proposed
multiclass conditional probability estimation method,
measured by
$$
\|\hat{p}_\lambda-p \|_1 =\sum_{k=1}^K \| \hat{p}_k-p_k \|_1 =
\sum_{k=1}^K E|\hat{p}_k({\bf X})-p_k({\bf X})|.
$$
The convergence rate is quantified in terms of the
tuning parameter $\lambda$, the number of brackets $m$, sample size
$n$, and the cardinality of ${\cal F}$.

\subsection{Asymptotic theory}

The following technical assumptions are made.

{\it Assumption 1.} For any $\tau \in (0,1)$, there exists
$\bar{f}_\tau \in {\cal F}$, such that $e(\bar{f}_\tau,
\tilde{f}^*_\tau)\leq s_n$ for some positive sequence $s_n
\rightarrow 0$ as $n \rightarrow \infty$.

This is analogous to Assumption 1 in Wang et al. (2008) and ensures
that the true quantile regression function $\tilde{f}^*_\tau$ can be well
approximated by ${\cal F}$.

{\it Assumption 2.} For any $\tau \in (0,1)$ and $f\in {\cal F}$, there exist constants $a_1>0$ and $0< \alpha \leq 1$ such that
$$
(e(f, \tilde{f}_\tau^*))^{\alpha} \geq a_1 \|f-\tilde{f}^*_\tau\|_1.
$$

Assumption 2 describes the local smoothness
of $f({\bf x})$ within the neighborhood of $\tilde{f}^*_\tau({\bf
x})$. Note that $e(f, \tilde{f}^*_\tau)=E( h_\tau({\bf X},
\tilde{Y}))$ with
$$
h_\tau({\bf x},y)=I(\tilde{f}^*_\tau({\bf x})\leq y \leq
f({\bf x}))(f({\bf x})-y)+I(f({\bf x})\leq y \leq
\tilde{f}^*_\tau({\bf x}))(y-f({\bf x}))
$$
by Lemma 4 in Li et al. (2007), so Assumption 2 is the same as Assumption A in Li et al. (2007).

Next we measures the cardinality of ${\cal F}$ by the $L_2$-metric
entropy with bracketing. Given any $\epsilon>0$, $\{(f_a^l, f_a^u),
a=1,\ldots,A\}$ is an $\epsilon$-bracketing function set of ${\cal
F}$ if for any $f\in {\cal F}$ there exists an $a$ such that
$f_a^l\leq f \leq f_a^u$, and $\|f_a^l-f_a^u \|_2 \leq \epsilon$ for
all $a=1,\ldots,A$. The $L_2$-metric entropy with bracketing
$H_B(\epsilon,{\cal F})$ is then defined as the logarithm of the
cardinality of the smallest $\epsilon$-bracketing function set of
${\cal F}$. Denote ${\cal F}(k)=\{f\in {\cal F}: J(f)\leq k \}$,
${\cal F}_\infty=\{f\in {\cal F}: J(f)< \infty \}$ and
$J_0=\min\limits_\tau \max\{J(\bar{f}_\tau),1\}$.

{\it Assumption 3.} For some positive constants $a_2,a_3$ and $a_4$,
there exists some $\epsilon_n >0$ such that
\begin{equation}\label{assum3}
\sup_{k \geq 1}~\phi (\epsilon_n,k) \leq a_2 n^{1/2},
\end{equation}
where $\phi (\epsilon_n,k)=\frac{1}{D}\int_{a_4
D}^{a_3^{1/2}D^{\alpha/2}} H_B^{1/2} (u,{\cal F}(k)) du$ and
$D=D(\epsilon_n,\lambda,k)=\min\{\epsilon_n^2+ (k-1) \lambda J_0,
1\}$.

{\theorem Suppose Assumptions 1-3 are met, and there
exists $T>0$ such that $\rho_\tau (y-f({\bf x})) \leq T$ for any $f \in {\cal F}$. For
$\hat{p}_{\lambda}({\bf x})$ obtained as in (\ref{scale1}),
\begin{equation}
Pr\left( \|\hat{p}_\lambda-p\|_1 \geq \frac{4K}{m}+ 2 K m^2 a_1^{-1}
\delta_n^{2 \alpha}\right) \leq  7 m K \exp\left(-a_5 n(\lambda
J_0)^{2-\alpha}\right), \label{bound}
\end{equation}
provided that $\lambda J_0 \leq \delta_n^2/2$, where
$\delta_n^2=\min\{\max(\epsilon_n^2,s_n),1\}$.}

{\corollary Under the assumptions in Theorem 1,
\begin{equation*}
\|\hat{p}_\lambda-p\|_1 = O_p \left(\frac{2}{m}+ m^2 a_1^{-1}
\delta_n^{2 \alpha}\right), ~~ E\|\hat{p}_\lambda-p\|_1 = O
\left(\frac{2}{m}+ m^2 a_1^{-1} \delta_n^{2 \alpha}\right),
\end{equation*}
provided that $n(\lambda J_0)^{2-\alpha}-\log(m)$ diverges as $n \rightarrow \infty$.}

Theorem 1 and Corollary 1 provide probability and risk bounds for
$\|\hat{p}_\lambda-p\|_1$. They also suggest the ideal $m$ to be of
order $O(\delta_n^{-2\alpha/3})$, yielding the fast rate of
$O_p(\delta_n^{2\alpha/3})$ for $\|\hat{p}_\lambda-p\|_1$.

\subsection{A theoretic example}

To illustrate the asymptotic theory, a simple theoretic example is
considered. Let ${\bf X}$ be sampled from a uniform distribution on
$(0,3)$ and $Y \in \{1, 2, 3\}$ be sampled according to $p_k(x)=0.8$
if $k-1 \leq x < k$ and 0.1 otherwise. Let ${\cal F}_1=\{f: f \in
{\cal H}_K, f(x) \in (0.5,3.5)\}$, where $K$ is the Gaussian kernel.

To verify Assumption 1, note that for any $\tau \in (0,1)$,
$\tilde{f}^*_\tau(x)$ is continuous in $x$ except at $x=1$ and
$x=2$. For given $s_n$, define
\[ g_\tau(x)= \left\{ \begin{array}{ll}
\tilde{f}_\tau^*(k-\frac{s_n}{8})+
\frac{x-(k-\frac{s_n}{8})}{\frac{s_n}{4}} (\tilde{f}_\tau^*(k+\frac{s_n}{8})-
\tilde{f}_\tau^*(k-\frac{s_n}{8})),
&\ \text{if}\ x \in (k-\frac{s_n}{8},k+\frac{s_n}{8}), k=1,2 ;\\
\tilde{f}_\tau^*(x),& \text{otherwise},
\end{array} \right. \]
then $g_{\tau}(x)$ is a continuous function of $x$, and $\|g_\tau-\tilde{f}^*_\tau\|_1 \leq s_n/2$. Furthermore, as
$g_\tau(x)$ is continuous, Steinwart (2001) shows that there exists a $\bar{f}_\tau \in {\cal F}_1$ such that $\|g_\tau-\bar{f}_\tau\|_1 \leq \|g_\tau-\bar{f}_\tau\|_\infty \leq s_n/2$. Therefore,
$\|\bar{f}_\tau-\tilde{f}^*_\tau\|_1 \leq \|g_\tau-
\tilde{f}^*_\tau\|_1+\|g_\tau-\bar{f}_\tau\|_1 \leq s_n$. Since
$|\rho_\tau (y-\bar{f}_\tau(x))-\rho_\tau
(y-\tilde{f}_\tau^*(x))| \leq |(y-\bar{f}_\tau(
x))-(y-\tilde{f}_\tau^*(x))| = |\bar{f}_\tau(
x)-\tilde{f}_\tau^*(x)|, $ then
\begin{equation*}
e(\bar{f}_\tau, \tilde{f}_\tau^*) =
E\Big (\rho_\tau(\tilde Y-\bar{f}_\tau(X))-\rho_\tau(\tilde Y-\tilde{f}_\tau^*(X)) \Big ) \leq E
\Big | \bar{f}_\tau(X)- \tilde{f}_\tau^*(X) \Big | =
\|\bar{f}_\tau - \tilde{f}_\tau^*\|_1\leq s_n.
\end{equation*}

To verify Assumption 2, note that $e(f, \tilde{f}_\tau^*) =
E( h_\tau (X, \tilde{Y}) ) =E\Big( E (h_\tau (
X, \tilde{Y})|X )\Big)$, and
\begin{eqnarray*}
& & E(h_\tau (X, \tilde{Y})|X) \\
&=& E\left(I(\tilde{f}^*_\tau(X)\leq y \leq f(X))(f(X)-\tilde Y)+I(f(X)\leq \tilde Y
\leq \tilde{f}^*_\tau(X))(\tilde Y-f(X))|X \right) \\
&=& \left| \int_{\tilde{f}^*_\tau(X)}^{f(X)} P_X(u)(f(X)-u)du \right|
\geq 0.1 \left|\int_{\tilde{f}^*_\tau(X)}^{f(X)}(f (X)-u)du \right|
= 0.05 |\tilde{f}^*_\tau(X)-f(X) |^2,
\end{eqnarray*}
where $P_X(u)=p_k(X)$ if $k-0.5 \leq u < k+0.5.$
Therefore, Assumption 2 is satisfied with $\alpha=0.5$ and
$a_1=\sqrt{0.05}$.

To verify Assumption 3, since ${\cal H}_B(u, {\cal
F}(k))=O(\log^2(k/u))$ (Zhou, 2002) for any given $k$ and $\phi
(\epsilon_n,k)$ is nonincreasing in $D$, there exist positive
constants $c_1 , c_2$, such that
$$
\sup_{k \geq 1}\phi(\epsilon_n,k) \leq \phi(\epsilon_n,1) = \frac{1}{D}\int_{a_4
D}^{a_3^{1/2}D^{\alpha/2}} c_1 \log(1/u) du \leq c_2
\log(1/\epsilon_n)/\epsilon_n^{2-\alpha}.
$$
Without loss of generality, assume $s_n \leq \epsilon_n^2 \leq 1$, and then $\delta_n^2=\epsilon_n^2$. Solving (\ref{assum3}), yields that
$\delta_n^2=O \Big ((\frac{\log^2 n}{n})^{1/(2-\alpha)} \Big)$, when $\lambda J_0 \sim \delta_n^2$.

Finally, by Corollary 1, $E\|\hat{p}_\lambda-p\|_1 = O
\left(\frac{2}{m}+ m^2 a_1^{-1} n^{-1/3} (\log n)^{2/3}\right)$.
This implies that $E\|\hat{p}_\lambda-p\|_1 = O\big(n^{-1/9}(\log
n)^{2/9}\big)$ when $m$ is set as $O \big ( n^{1/9} (\log n)^{-2/9}\big )$.

\section{Numerical experiments}

This section examines the effectiveness of the proposed multiclass probability estimation method in simulated and real examples. The numerical performance of the proposed method (OUR) is compared against three popular competitors: baseline logistic model (BLM), classification tree (TREE) and weighted multiclass classification (WMC; Wu et al., 2010). For illustration, the number of quantiles $m$ in our method is set as $m=100$. The kernel function used in each method is set as the Gaussian kernel $K({\bf z}_1,{\bf z}_2)=e^{-\|{\bf z}_1-{\bf z}_2\|^2/2 \sigma^2}$, where the scale parameter $\sigma^2$ is set as the median of pairwise Euclidean distances within the training set. To optimize the performance of each estimation method, a grid search is employed to select the tuning parameter as in Section 2.2. The grid used in all examples is set as $\{10^{(s-31)/10}; s=1,\ldots,61\}$. A more refined grid search can be employed to further improve the numerical performance at the cost of increased computation burden.

In simulated examples where the true conditional probability $p_k({\bf x})$ is known, the performance of each estimation method is measured by its distance to $p_k({\bf x})$. Various distance measures between $\hat p_k({\bf x})$ and $p_k({\bf x})$ are computed based on the testing set,
\begin{eqnarray*}
\mbox{1-Norm error:} & & err_1(\hat{p}_\lambda, p)=\frac{1}{|T|} \sum_{t \in T} \sum_{k=1}^K |\hat{p}_k({\bf x}_t)-p_k ({\bf x}_t)|; \\
\mbox{2-Norm error:} & & err_2(\hat{p}_\lambda, p)=\frac{1}{|T|} \sum_{t \in T} \sum_{k=1}^K (\hat{p}_k({\bf x}_t)-p_k ({\bf x}_t))^2; \\
\mbox{GKL loss:} & & err_{KL}(\hat{p}_\lambda, p)=\frac{1}{|T|} \sum_{t \in T}
\sum_{k=1}^K p_k ({\bf x}_t) \text{log}\frac{p_k({\bf x}_t)}{\hat{p}_k({\bf x}_t)}; \\
\mbox{Cross entropy error (CEE):} & & err_{CE}(\hat{p}_\lambda)=\frac{1}{|T|} \sum_{t \in T}  -\text{log} \left(\hat{p}_{y_t} ({\bf x}_t)\right),
\end{eqnarray*}
where $T$ denotes the testing set, and $|T|$ is the cardinality of $T$. To avoid degeneration in computing GKL loss and CEE, a small correction constant $0.01$ is added to $\hat{p}_k({\bf x})$ when necessary.

\subsection{Simulated examples}

Five simulated examples are generated for comparison.

{\it Example 1.} First, $Y$ is generated uniformly over  $\{1,2,3,4,5\}$. Next, given $Y=y$, the covariates ${\bf X}$ are generated from $\text{T}(\mu(y),\Sigma,df=2)$, a multivariate $t$ distribution with $\mu(y)=(\cos(2y\pi/5),\sin(2y\pi/5))^T$, $\Sigma= diag(1,2)$ and degree of freedom 2. The training size is 400, and the testing size is 2600.

{\it Example 2.} First, $Y$ is generated uniformly over  $\{1,2,\ldots,10\}$. Next, given $Y=y$, the covariates ${\bf X}$ are generated from $\text{T}(\mu(y),\Sigma,df=2)$, where  $\mu(y)=(\cos(y\pi/5),\sin(y\pi/5))^T$ and $\Sigma= diag(1,2)$. The training size is 400, and the testing size is 2600.

{\it Example 3.} First, $Y$ is generated uniformly over  $\{1,2,\ldots,20\}$. Next, given $Y=y$, the covariates ${\bf X}$ are generated from $\text{T}(\mu(y),\Sigma,df=2)$, where  $\mu(y)=(\cos(y\pi/10),\sin(y\pi/10))^T$ and $\Sigma= diag(1,2)$. The training size is 400, and the testing size is 2600.

{\it Example 4.} First, $Y$ is generated uniformly over  $\{1,2,3,4,5\}$. Next, given $Y=y$, the covariates ${\bf X}$ are generated from $\text{T}(\mu(y),\Sigma,df=2)$, where  $\mu(y)=(\cos(2y\pi/5),\sin(2y\pi/5),$ $0,\ldots,0)^T$, and $\Sigma= diag(1,2,1,\ldots,1)$ if $y$ is odd and $\Sigma= diag(2,1,1,\ldots,1)$ if $y$ is even. The training size is 400, and the testing size is 2600.

{\it Example 5.} First, $Y$ is generated uniformly over  $\{1,2,\ldots,10\}$. Next, given $Y=y$, the covariates ${\bf X}$ are generated from $\text{T}(\mu(y),\Sigma,df=2)$, where $\mu(y)=(\cos(y\pi/5),\sin(y\pi/5),$ $0,\ldots,0)^T$, and $\Sigma= diag(1,2,1,\ldots,1)$ if $y$ is odd and $\Sigma= diag(2,1,1,\ldots,1)$ if $y$ is even. The training size is 400, and the testing size is 2600.

Examples 1-3 are generated similarly, but with different number of classes $K$ and different mean vectors $\mu(y)$. When $K$ gets larger, the generated data from different classes become more overlapped and thus the resultant classification becomes more difficult. Examples 4 and 5 include additional noise variables and heteroscedastic covariance matrices. Each simulated example is repeated 50 times, and the averaged test errors and the corresponding standard deviations are reported in Table 1.

\begin{center}
\begin{tabular}{c}
\hline
\hline
Table 1 here\\
\hline
\hline
\end{tabular}
\end{center}

Evidently, the proposed estimation method delivers superior numerical performance, and outperforms BLM, TREE and WMC in all the examples. As a model-free method, WMC yields competitive performance in Example 1 and Example 4 with $K=5$ where the data from different classes are relatively far apart leading to clear-cut classification boundary. However, when $K$ gets larger, WMC requires much more intensive computing power, and its numerical performance appears to be less satisfactory in Examples 2 and 5 with $K=10$. Furthermore, the performance of WMC in Example 3 with $K=20$ is not reported in Table 1, since it is computationally expensive to achieve reasonably good estimation accuracy.

\subsection{Real applications}

In this section, the proposed multiclass probability
estimation method is applied to the iris data, the white wine quality data and the abalone data. All datasets are publicly available at the University of California Irvine Machine Learning Repository ({\it http://archive.ics.uci.edu/ml/}).

The iris data has 4 continuous attributes: sepal length, sepal width, petal length, and petal width, and three classes: Setosa, Versicolour, and Virginica. The size of the iris dataset is 150, and each class has 50 observations. We randomly select 30 observations from each class and set as the training set, and the remaining 60 observations are used for testing. The white wine quality data has 11 attributes, which characterize various aspects of the white wines, and the response ranges from 0 to 10 representing quality scores made by wine experts. For illustration, we focus only on three classes with quality scores 5, 6 and 7, and a total of 4535 white wines are selected, where 1457, 2198 and 880 white wines score 5, 6 and 7, respectively. We randomly select 100 white wines from each class as the training set, and the remaining 4235 white wines are used for testing. The abalone data has 8 attributes on various physical measurements of an abalone, and 29 classes representing different ages of an abalone. Since some extreme classes have very few abalones, we only focus on the $K$ largest classes with $K=5, 8, 10$. In specific, for $K = 5$, classes $7 - 11$ are selected with a total of 2768 abalones; for $K = 8$, classes $6 - 13$ are selected with a total of 3498 abalones; for $K = 10$, classes $5 - 14$ are selected with a total of 3739 abalones. In all scenarios, we randomly select 50 abalones from each class as the training set, and keep the remaining abalones for testing.     

Note that the true conditional probability $p_k({\bf x})$ is not available in the real applications, so only CEE is computed and used for comparison. In addition, we also compare the averaged misclassification error (MCE) of each probability estimation method on the testing set, where the classification label is predicted as $\hat y_t = \argmax_k \hat p_k({\bf x}_t)$, and MCE is defined as
$$
MCE(\hat{p}_\lambda, p) =  \frac{1}{|T|} \sum_{t \in T}  I(\hat y_t \neq y_t).
$$
The averaged CEE and MCE over 50 replications are reported in Table 2.

\begin{center}
\begin{tabular}{c}
\hline
\hline
Table 2 here\\
\hline
\hline
\end{tabular}
\end{center}

It is evident that the proposed probability estimation method delivers competitive results against other competitors. It yields the smallest CEE and MCE in all real examples, except that WMC produces slightly smaller CEE in the iris example. The performance of WMC is not reported for the abalone example with $K=8$ and $10$ due to the computational burden.

\section{Summary}

This paper proposes an efficient model-free multiclass conditional probability
estimation method, where the estimated probabilities are constructed via a series of estimated conditional quantile regression functions.
The proposed method does not require any distributional model assumption, and it is computationally efficient as its computation cost does not need to increase exponentially with $K$.
The asymptotic convergence rate of the proposed method is established, and the numerical experiments with both simulated examples and real applications demonstrate the advantage of the proposed method, especially when $K$ is large. In addition, $p_k({\bf x})=P(Y=k|{\bf x})$ can be regarded as the conditional density of discrete $Y$, and thus the proposed method can be naturally extended to a general framework of conditional density estimation (Hansen, 2004).

\section*{Appendix: technical proofs}

{\bf Proof of Lemma 1.} When $\sum\limits_{j=1}^{k-1} p_j({\bf x})<
\tau \leq \sum\limits_{j=1}^k p_j({\bf x})$,
\begin{eqnarray*}
P(\tilde{Y}\leq \tilde{f}^*_\tau ({\bf x})) &=& P(Y \leq
k-1)+Pr \Big (Y=k, -0.5\leq \varepsilon \leq
\frac{\tau-\sum\limits_{j=1}^{k-1} p_j({\bf x})} {p_k({\bf x})}-0.5 \Big ) \\
&=& \sum\limits_{j=1}^{k-1} p_j ({\bf x})+p_k(x)\times
\frac{\tau-\sum\limits_{j=1}^{k-1} p_j({\bf x})} {p_k({\bf
x})}=\tau.
\end{eqnarray*}
The desired result follows immediately.

\noindent{\bf Proof of Theorem 1.}
First, note that $p_k({\bf x}) = \sum\limits_{s=0}^{k} p_s({\bf
x})-\sum\limits_{s=0}^{k-1} p_s({\bf x})$ with $p_0({\bf x})=0$, and then
\begin{eqnarray}
\|\hat{p}_k-p_k\|_1&=&\left\|\sum\limits_{s=0}^{k}
\hat{p}_s-\sum\limits_{s=0}^{k-1} \hat{p}_s-\sum\limits_{s=0}^{k}
p_s+\sum\limits_{s=0}^{k-1} p_s \right\|_1 \nonumber \\
&\leq&
\left\|\sum\limits_{s=0}^{k}\hat{p}_s-\sum\limits_{s=0}^{k}p_s
\right\|_1+
\left\|\sum\limits_{s=0}^{k-1}\hat{p}_s-\sum\limits_{s=0}^{k-1}p_s\right\|_1. \label{scalethm1}
\end{eqnarray}
Therefore, it suffices to bound $\left\|\sum\limits_{s=0}^{k}\hat{p}_s-\sum\limits_{s=0}^{k}p_s
\right\|_1$ for any $k$.

Next, for simplicity, denote $\hat{P}_k({\bf
x})=\sum\limits_{s=0}^k \hat{p}_s({\bf x})$, $P_k({\bf
x})=\sum\limits_{s=0}^k p_s({\bf x})$, and $B_k=\Big \{{\bf x}: \Big |\hat{P}_k({\bf x})- P_k({\bf x}) \Big |\geq \frac{2}{m} \Big \}$. Simple calculation yields that
\begin{eqnarray*}
\|\hat{P}_k-P_k\|_1 &=& E|\hat{P}_k({\bf X}) - P_k({\bf X})| \\
&=&E\left( |\hat{P}_k({\bf X}) - P_k({\bf X}) | \cdot
I(B_k)\right)+E\left(|\hat{P}_k({\bf X}) - P_k({\bf X})|\cdot I(B_k^c)\right)\\
&\leq& P(B_k)+\frac{2}{m} P(B_k^c)\leq P(B_k)+\frac{2}{m},
\end{eqnarray*}
where the first inequality follows from the fact that $|\hat{P}_k({\bf X}) - P_k({\bf X}) |$ is bounded by 1. Therefore, bounding $\|\hat{P}_k-P_k\|_1$ boils down to bounding $P(B_k)$.

Based on the estimation method in (\ref{scale1}), there exists $j_1 \in \{1,\ldots,m-1\}$, such that $\tau_{j_1}=\hat{P}_k({\bf x})$, and then
$\hat{f}_{\tau_{j_1}}({\bf x}) \leq k+0.5$ and $\hat{f}_{\tau_{j_1+1}}({\bf x}) > k+0.5$. Let $\triangle_j=\{{\bf x}:
|\hat{f}_{\tau_j}({\bf x})-\tilde{f}^*_{\tau_j}({\bf x})|\geq \frac{1}{m}\}$, and we will show the relationship $B_k= \Big \{{\bf x}: |\hat{P}_k({\bf
x})-P_k({\bf x}) | \geq \frac{2}{m} \Big \} \subset \bigcup_{j=1}^{m-1}\triangle_j$ in the following four cases.

{\it Case 1.}  If $\hat{P}_k({\bf x})-P_k({\bf x})\geq \frac{2}{m}$ and $\hat{P}_k({\bf x}) \leq P_{k+1}({\bf x})$, then $P_k({\bf x}) + \frac{2}{m} \leq \tau_{j_1}=\hat{P}_k({\bf x}) \leq P_{k+1}({\bf x})$. Based on Lemma 1,
$$
\tilde{f}^*_{\tau_{j_1}}({\bf x}) = k+0.5+\frac{\hat{P}_k({\bf x})-P_k({\bf x})}{p_{k+1}({\bf x})}
\geq k+0.5+\hat{P}_k({\bf x})-P_k({\bf x})\geq k+0.5+\frac{2}{m},
$$
which implies that $\tilde{f}^*_{\tau_{j_1}}({\bf x})-\hat{f}_{\tau_{j_1}}({\bf x}) \geq \frac{2}{m} > \frac{1}{m}$.

{\it Case 2.}  If $\hat{P}_k({\bf x})-P_k({\bf x})\geq \frac{2}{m}$ and $\hat{P}_k({\bf x}) >  P_{k+1}({\bf x})$, then by Lemma 1, $\tilde{f}^*_{\tau_{j_1}}({\bf x}) > k+1.5$ and $\tilde{f}^*_{\tau_{j_1}}({\bf x})-\hat{f}_{\tau_{j_1}}({\bf x}) > 1 > \frac{1}{m}$.

{\it Case 3.} If $P_k({\bf x})-\hat{P}_k({\bf x}) \geq \frac{2}{m}$ and $\hat{P}_k({\bf x}) >  P_{k-1}({\bf x})-\frac{1}{m}$, then $P_{k-1}({\bf x}) - \frac{1}{m} < \tau_{j_1} \leq P_k({\bf x}) - \frac{2}{m} $ and $P_{k-1}({\bf x}) < \tau_{j_1+1} = \tau_{j_1} +\frac{1}{m} \leq P_k({\bf x}) - \frac{1}{m} $. Based on Lemma 1,
$$
\tilde{f}^*_{\tau_{j_1+1}}({\bf x}) = k - 0.5 + \frac{\tau_{j_1+1}-P_{k-1}({\bf x})}{p_k({\bf x})}
\leq k - 0.5 + \frac{p_{k}({\bf x})-\frac{1}{m}}{p_k({\bf x})} \leq k+0.5 - \frac{1}{m},
$$
which implies that $\hat{f}_{\tau_{j_1+1}} -\tilde{f}^*_{\tau_{j_1+1}} >
\frac{1}{m}$.

{\it Case 4.} If $P_k({\bf x})-\hat{P}_k({\bf x}) \geq \frac{2}{m}$ and $\hat{P}_k({\bf x}) \leq  P_{k-1}({\bf x})- \frac{1}{m}$, then $\tau_{j_1+1}\leq P_{k-1}({\bf x})$ and by Lemma 1, $\tilde{f}^*_{\tau_{j_1+1}}({\bf x}) \leq k-0.5$ and $\hat{f}_{\tau_{j_1+1}}({\bf x})-\tilde{f}^*_{\tau_{j_1+1}}({\bf x}) > 1 > \frac{1}{m}$.

Combining the above four cases, $B_k \subset \bigcup_{j=1}^{m-1}\triangle_j = \{ {\bf x}: |\hat{f}_{\tau_j}({\bf
x})-\tilde{f}^*_{\tau_j}({\bf x})|\geq \frac{1}{m}~\text{for some}~j \}$. It leads to a connection between
$\|\hat{P}_k-P_k \|_1$ and
$e(\hat{f}_{\tau},\tilde{f}^*_{\tau}) $ is established in the following.
\begin{eqnarray*}
& & \left\{\|\hat{P}_k-P_k\|_1 \geq \frac{2}{m}+m^2 a_1^{-1}
\delta_n^{2 \alpha}\right\} \subset \Big\{P(B_k)\geq m^2 a_1^{-1}\delta_n^{2\alpha} \Big\} \\
&\subset& \Big\{Pr\Big(\bigcup_{j=1}^{m-1} \triangle_j \Big)\geq
m^2 a_1^{-1} \delta_n^{2\alpha}\Big\} \subset \Big\{P(\triangle_j)\geq m a_1^{-1}\delta_n^{2\alpha},\ \text{for
some j} \Big\}.
\end{eqnarray*}
Therefore, $Pr\left(\|\hat{P}_k-P_k\|_1 \geq
\frac{2}{m}+m^2 a_1^{-1} \delta_n^{2 \alpha}\right) \leq
\sum\limits_{j=1}^{m-1} Pr\left(P(\triangle_j) \geq m
a_1^{-1}\delta_n^{2\alpha}\right)$.
In addition, $P(\triangle_j)\geq m a_1^{-1} \delta_n^{2\alpha}$
implies that $\|\hat{f}_{\tau_j}-\tilde{f}^*_{\tau_j}\|_1 \geq
\frac{1}{m}P(\triangle_j)=a_1^{-1} \delta_n^{2\alpha}$. This,
together with Assumption 2, yields that
$e(\hat{f}_{\tau_j},\tilde{f}^*_{\tau_j})\geq \delta_n^2$.
Therefore,
\begin{eqnarray*}
& & Pr\left( \|\hat{P}_k-P_k\|_1 \geq \frac{2}{m}+m^2 a_1^{-1}
\delta_n^{2 \alpha}\right) \leq \sum\limits_{j=1}^{m-1}
Pr\left(P(\triangle_j) \geq
m a_1^{-1}\delta_n^{2\alpha}\right)\\
&\leq& \sum\limits_{j=1}^{m-1}
Pr \Big (e(\hat{f}_{\tau_j},\tilde{f}^*_{\tau_j})\geq \delta_n^2 \Big ) \leq m\cdot \max\limits_j \Big \{3.5\exp(-a_5 n(\lambda
J_{\tau_j})^{2-\alpha}) \Big \}\\
&\leq& 3.5 m \exp(-a_5 n(\lambda J_0)^{2-\alpha}),
\end{eqnarray*}
where the second to the last inequality follow from a slightly modified version of Theorem 2 in Li et al. (2007) incorporating the approximation error in Assumption 1.
\par
Based on (\ref{scalethm1}), $\|\hat{p}_k-p_k\|_1 \geq \frac{4}{m}+2 m^2 a_1^{-1} \delta_n^{2\alpha}$ implies that at least one of $\|\hat{P}_k-P_k\|_1$ and $\|\hat{P}_{k-1}-P_{k-1}\|_1$
is larger than $\frac{2}{m}+m^2 a_1^{-1} \delta_n^{2\alpha}$. Therefore,
\begin{eqnarray*}
& & Pr\left(\|\hat{p}_\lambda-p\|_1 \geq \frac{4K}{m}+ 2 K
m^2 a_1^{-1} \delta_n^{2 \alpha}\right) \\
&\leq& Pr\left(\bigcup_{k=1}^K
\left\{\|\hat{p}_k-p_k\|_1 \geq \frac{4}{m}+ 2 m^2 a_1^{-1} \delta_n^{2
\alpha}\right\}\right)  \leq \sum\limits_{k=1}^K Pr\Big(\|\hat{p}_k-p_k\|_1 \geq \frac{4}{m}+ 2 m^2 a_1^{-1} \delta_n^{2
\alpha}\Big ) \\
&\leq & \sum_{k=1}^K Pr\Big(\|\hat{P}_k-P_k\|_1 \geq \frac{2}{m}+ m^2 a_1^{-1} \delta_n^{2
\alpha}\Big) +  \sum_{k=1}^K Pr\Big(\|\hat{P}_{k-1}-P_{k-1}\|_1 \geq \frac{2}{m}+ m^2 a_1^{-1} \delta_n^{2
\alpha}\Big) \\
&\leq& 7 m K \exp(-a_5 n(\lambda J_0)^{2-\alpha}).
\end{eqnarray*}

{}

\newpage
\begin{table}[!h]
      \begin{center}
      \caption{Simulated examples. Estimated means and standard deviations (in parentheses) of 1-norm, 2-norm, GKL loss and CEE for various estimation methods based on 50 replications.}
      \bigskip
        \begin{tabular}{lrrrr}
         \hline
         \hline
                    &1-norm~~~~ & 2-norm~~~~ & EGKL~~~~ & CEE~~~~~~ \\
              \hline
              \textit{Example 1}\\
              \hline
              OUR & 0.316(0.0151)& 0.034(0.0034) & 0.080(0.0076) & 1.426(0.0122)\\
              BLM & 0.447(0.0357)& 0.068(0.0082) & 0.211(0.0256) & 1.552(0.0329)\\
              TREE & 0.404(0.0293)& 0.061(0.0091) & 0.178(0.0370) & 1.521(0.0403)\\
              WMC   & 0.336(0.0363)& 0.040(0.0089) & 0.139(0.0283) & 1.483(0.0354)\\
              \hline
              \textit{Example 2}\\
              \hline
              OUR & 0.359(0.0179)& 0.023(0.0026)& 0.104(0.0108)& 2.143(0.0168)\\
              BLM & 0.468(0.0345)& 0.042(0.0048)& 0.245(0.0325)& 2.279(0.0406)\\
              TREE & 0.511(0.0396)& 0.051(0.0095)& 0.256(0.0480)& 2.289(0.0514)\\
              WMC   & 0.560(0.0547)& 0.063(0.0151)& 0.270(0.0501)& 2.305(0.0597)\\
              \hline
              \textit{Example 3}\\
              \hline
              OUR & 0.412(0.0191)& 0.015(0.0016)& 0.136(0.0146)& 2.865(0.0166)\\
              BLM & 0.505(0.0286)& 0.027(0.0030)& 0.307(0.0460)& 3.032(0.0542)\\
              TREE & 0.714(0.0469)& 0.056(0.0084)& 0.352(0.0476)& 3.085(0.0527)\\
              WMC  & $--$~~~~~~~  & $--$~~~~~~~ & $--$~~~~~~~ & $--$~~~~~~~\\
              \hline
              \textit{Example 4}\\
              \hline
              OUR & 0.442(0.0163)& 0.072(0.0041)& 0.167(0.0100)& 1.496(0.0140)\\
              BLM & 0.555(0.0194)& 0.114(0.0070)& 0.345(0.0324)& 1.676(0.0344)\\
              TREE &0.679(0.0539)& 0.169(0.0249)& 0.568(0.0855)& 1.901(0.0859)\\
              WMC  &0.481(0.0510)& 0.084(0.0181)& 0.231(0.0468)& 1.561(0.0498)\\
          \hline
          \textit{Example 5}\\
          \hline
              OUR &0.486(0.0188)&0.044(0.0032)&0.201(0.0144)&2.217(0.0153)\\
              BLM &0.681(0.0209)&0.098(0.0076)&0.552(0.0558)&2.573(0.0581)\\
              TREE &0.790(0.0556)&0.124(0.0143)&0.612(0.0807)&2.621(0.0861)\\
              WMC  &0.702(0.0584)&0.095(0.0193)&0.455(0.0889)&2.472(0.0913)\\
          \hline
          \hline
        \end{tabular}
        \label{table1}
       \end{center}
\end{table}

\begin{table}[!h]
      \begin{center}
      \caption{Real applications. Estimated means and standard deviations (in parentheses) of CEE and MCE for various estimation methods base on 50 replications.}
      \bigskip
     \begin{threeparttable}
       \begin{tabular}{l r r}
         \hline
         \hline
                    &CEE~~~~~~   & MCE~~~~~~  \\
         \hline
          \textit{Iris example}\\
         \hline
              OUR   &0.167(0.0313) &0.041(0.0224) \\
              BLM   &2.164(0.0190) &0.051(0.0260) \\
              TREE   &0.220(0.0851)  &0.066(0.0234)  \\
              WMC    &0.146(0.0505)   &0.052(0.0245)   \\
          \hline
           \textit{Wine quality example}\\
          \hline
              OUR   & 0.926(0.0125)      & 0.468(0.0132) \\
              BLM   & 0.983(0.0227)      & 0.510(0.0133) \\
              TREE  & 1.472(0.1215)      & 0.530(0.0230) \\
              WMC   & 0.945(0.0226)      & 0.490(0.0241) \\
          \hline
           \textit{Abalone example $K=5$}\\
          \hline
              OUR   & 1.391(0.0121)      & 0.640(0.0138)  \\
              BLM   & 1.495(0.0483)      & 0.661(0.0101) \\
              TREE  & 1.975(0.0694)      & 0.732(0.0270) \\
              WMC   & 1.930(0.1440)      & 0.675(0.0190) \\
          \hline
           \textit{Abalone example $K=8$}\\
           \hline
              OUR   & 1.742(0.0123)      &  0.721(0.0120)\\
              BLM   & 2.173(0.2202)      &  0.775(0.0174)\\
              TREE  & 2.040(0.0834)      &  0.749(0.0208)\\
              WMC   & $--$~~~~~~~  		 & $--$~~~~~~~\\
            \hline
            \textit{Abalone example $K=10$}\\
           \hline
              OUR   & 1.910(0.0203)      & 0.756(0.0146) \\
              BLM   & 3.050(0.4905)      & 0.834(0.0197) \\
              TREE  & 2.149(0.1130)      & 0.771(0.0291) \\
              WMC   & $--$~~~~~~~  		 & $--$~~~~~~~ \\
           \hline
          \hline
        \end{tabular}
        \label{table2}
        \end{threeparttable}
        \end{center}
\end{table}

\begin{figure}[!h]
\caption{A solution surface of $\hat{f}_{\lambda,\tau}$ as a function of $(\lambda,\tau)$ in a
randomly selected replication of Example 1.}
\begin{center}
\includegraphics[width=0.5\textwidth]{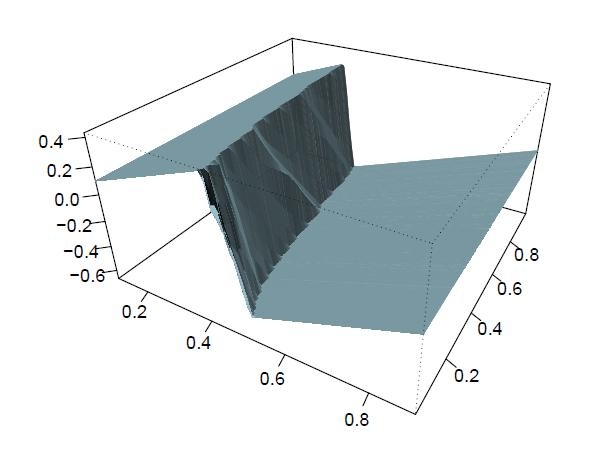}
\end{center}
\end{figure}

\end{document}